# A Hybrid Deep Learning Approach for Texture Analysis

Hussein Mohamed Adly
Department of Computer Science
American University in Cairo
Giza, Egypt
husseinadly@aucegypt.edu

Mohamed Moustafa
Department of Computer Science
American University in Cairo
Giza, Egypt
m.moustafa@aucegypt.edu

*Abstract*—**Texture classification is a problem that has various applications such as remote sensing and forest species recognition. Solutions tend to be custom fit to the dataset used but fails to generalize. The Convolutional Neural Network (CNN) in combination with Support Vector Machine (SVM) form a robust selection between powerful invariant feature extractor and accurate classifier. The fusion of experts provides stability in classification rates among different datasets.**

*Keywords; texture, deep learning, cnn, svm, confusion matrix, classifications, fusion*

## I. Introduction

Textures are slowly varying and almost periodically repeating patterns that compose a scene or image. Textures are not yet well understood in part due to difficulty in understanding human visual perception system regarding the type of visual clues used by humans to understand images[1]. This general property is commonly used for extraction of local textural features, i.e. some statistical characteristics that could be utilized for an adequate description of texture patterns that shall be used for comparing and matching textures. Texture analysis uses mathematically based models to describe spatial variations in images or scenes[2]since it's easier to describe textures regarding statistical patterns than geometrical edges. It is utilized in various fields including fire smoke detection[3], remote sensing, forest species recognition, scene segmentation, content-based image retrieval, industrial inspection, etc. Thus, texture classification problem remains to be one of key pattern recognition tasks. The primary concern of texture analysis is textural features extraction and accurate matching of such features.

Texture features are divided into three categories statistical, structural or geometrical and digital signal processing methods. There is although a great overlap and commonalities between statistical methods and digital signal processing methods due to using same mathematical tools for achieving their purpose. The statistical methods rely on the statistics of the spatial distribution of usually constant or slowly varying gray level values. The spatial statistics are used as a pattern to match it with other statistical information. An example of this method is the co-occurrence and the autocorrelation function. Geometrical methods, which decompose textures into simple geometrical primitives and their placement usually by using methods, like edge detection with a Laplacian-of-Gaussian. The simple geometrical methods fail to distinguish between edges, fine details, and neighborhoods[4]. Statistical methods could be used to describe patterns in the structural method of simple geometrical primitives. Signal processing methods analyze the frequency domain of spatial information. Fourier analysis is used to analyze frequency domain, and it's very effective when textures exhibit high periodicity, while autocorrelation shows the repetition of textures as well as local intensity variations [2].

Convolutional Neural Networks (CNN) since its inception had shown promising results, especially with computer vision problem, and as a result classifier such as SVM became less popular. The SVM in various image classification experiments had demonstrated near state of the art results among other classifiers which make it potentially a good complementary classifier supporting the CNN for better overall classification results. Multiple classifiers solutions were used successfully in more than an occasion to improve classification results.

## II. Previous Work

Various approaches were used to explore the problem of texture analysis and improve the classification results. Lei et al.[5], developed a new feature extractor Complete LBP (CLBP). An image gray level local region is represented by its center pixel, and global thresholding is done before binary coding the center pixel to generate rotational invariant features.

Khalid et al. [6] used Local Binary Patterns (LBP) for feature extraction and K nearest-neighbor (KNN) for classification. Tou et al. [7] proposed using co-occurrence matrices (GLCM) and Gabor filters. Paula et al. used color-based features and GLCM in [8], and used a mixture of feature extractors LBP, CLBP, GLCM and color features in [9]. Li Li et al. [10] proposed Color Local Gabor binary Co-occurrence Pattern (CLGBOCP) that finds the spatial relationship of a pixel's neighborhoods using co-occurrence local binary edges. CLGBOCP can integrate LBP with co-occurrence matrix features for complete feature extraction.

Liao et al.[11] proposed linear Dominant Local Binary Patterns (DLBP) that calculate the frequency of occurrence of rotational invariant patterns. The extracted patterns are sorted in descending order from the most dominant to least dominant. To extract global features, Circularly Symmetric Gabor Filter, which is rotation invariant, and less sensitive to histogram equalization than DLBP. The aggregation of the local and global features into a classifier yields better results than either of them could achieve independently.

Recently, researchers started adopting deep learning to solve the problem. Luiz Hafemann[12] proposed using convolutional deep learning architecture to determine features and analyze textures. The results were not consistently better than previously reported results.

A variant from the proposed methodology using the same types of classifiers was employed by [13] for object categorization and reported a decrease in error rate of 1.3% compared to the best standalone classifier. Similarly [14] used a variant fusion scheme of the same subset of classifiers, reporting the success of the combination at the recognition of hand gestures.

## III. SVM, CNN, AND DATASETS

### A. Convolutional Neural Network

CNNs are layered learning approach where higher layer learns higher level patterns than previous layers.In CNN, neurons are only connected to only a subset of the neurons of the previous layer, which is called receptive field. The chosen sub- set of neurons of the receptive field size is dynamic based on the filter size.The width, height, and depth of filters are treated in a different manner. For height and depth, the connections are local, and it goes along the full depth of receptive fields. The CNN is composed of Convolutional, Sub-sampling, ReLU, and fully connected layers. The convolutional layers mimic mammals visual system by utilizing filters to produce a new variant that focuses on pattern extraction from input data. The subsampling layer uses filters to resize the spatial dimensions of data with minimal data loss. As a result, the CNN becomes more robust to transformations and computationally more efficient. The fully connected layer acts as a converter of the feature map of the previous layer into a single feature vector. ReLU layers work as activation layers beyond the threshold of zero [14]. Figure 1 shows the architectural design of CNN.

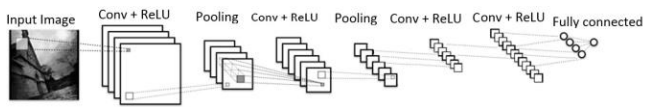

Fig. 1. Convolutional Neural Network Architecture

At the preprocessing phase, mean subtraction is applied to the training data and the means is saved to be used in testing and validation phases. Scaling initial weights produced by the random distribution of filters by 0.01 is applied to make sure the weights are not symmetric. Thus, there is no need for bias to be initialized by non-zero number because bias is used ensure weights have an asymmetrical shape for the gradient to work and ReLU shall be activated initially. The ReLU activation at initial learning stages isn't likely to introduce marginal performance increase. Initially, the first layer employs 64 filters with the depth of one and filters size 5x5, which is used to grasp the patterns on a very high level. For the CNN to handle variations, the network should capture complete patterns. A moderately sized filter size receptive field corresponding to input spatial size would allow the network to obtain whole patterns. Choosing the number of filters must be done with caution since a large number of filters may improve accuracy due to the stacking of filters to form receptive fields, but that will also increase the time and computational complexity while few of them usually leads to a lack of generalization. Convolutional layers have fixed stride to prevent mistakes in padding calculation, which is employed to preserve the output filter volume spatially. The filter depth should be proportional to a number of input filters from the previous layer while the filter number indicates how many filters with the aforementioned filter depth is used. The filter number is a hyperparameter and not enforced, unlike the depth. The pooling layer is used to reduce the computational parameters as the depth increase and provide non-linearity.

### B. Support Vector Machine

SVMs primary objective is finding the maximum marginal separation hyperplane of feature vectors. It is not computationally intensive nor prone to noisy data compared to neural networks and performs well in higher dimensional or attributes spaces. SVM is not highly prone to overfitting since generalization does not depend on dimensionality. It neither fall for local optimum [15]nor critically suffers from small training data since it works by structural risk minimization. SVM relies on Gamma and C for optimization where Gamma defines how wide the effect of a class of a support vector could have on other support vectors depending on the distance between the two. If Gamma is large, a support vector won't take part in classifying far vectors, and if it's too small, the whole training set will have an influence on the classification decision, and thus it's unlikely that a generalization would be reached. Gigantic values of Gamma will make the only affecting factor the support vector itself, and overfitting will be inevitable regardless of C. The C works as how tolerable are errors since allowing errors sometimes could lead to better convergence in cases where outliers exist. The higher the C, the higher the plenty of misclassification which may not usually good to give large plenty since the spreading of the data might be very far from linear distribution. The two widely used feature extractors are LBP and GLCM. The LBP suffers from rotational invarience[16]. Thus, SVM feature extraction is extracted using GLCM since it was shown that GLCM has a high rate in differentiating between intra-class variances[1]. The features are calculated with eight levels, three distances four directions of [0 1], [-1 1], [-1 0], [-1 -1] since the measure of co-occurrence obtained at angle 0 would be equal to 180, 45 equal to 225, etc. The distances were chosen based on empirical studies that had shown that values between 1 and

10 give better accuracy. The values around 1 and 2 had shown highest accuracy. We had chosen 1, 3 and 5 to capture patterns within large distance, yet not too large so that the GLCM wouldn't capture detailed texture information. Thirteen different features for each distance and direction is calculated in addition to the mean of the four directions at each distance. SVM is then trained with C of 2048 and gamma of 0.0313.

## C. Datasets

The Brodatz textures are de-facto standard and widely used as a benchmark dataset in texture segmentation and classifica- tion. It consists of 112 textures that were abstracted from the Brodatz texture album. Each of these textures is produced from a single image scanned from the texture album. Brodatz32 contains 32 texture classes where each one of the 32 classes has 64 samples, 16 of them are unique while others are variations of the unique 16. The variations are transformations such are rotation, scaling, or both.

Kylberg image data capture conditions are not ideal, so artificial corrections using photo editing tools were introduced by t h e authors. It has 28 classes where each of them has 160 samples of size 567x567. The images were resized to 64 x 64 to make it compatible with the CNN architecture designed to work with moderately sized images in dimension.

## IV. METHODOLOGY

### A. CNN-SVM Fusion

The CNN - SVM fusion exploits the invariant features learning of CNN and high accuracy of separation of feature vectors with SVM[13].

The Training set is used for training both classifiers CNN and SVM while the validation set is utilized only by CNN to check for an appropriate number of epochs before stopping the training and combat overfitting. The CNN was trained for 150 epochs for Brodatz dataset and 100 epochs for Kylberg dataset. The Test set is used to monitor the performance of each classifier for an individual class and assigns the appropriate classifier for that class given average statistics over a number of iterations defined empirically. Binary mapping is utilized to actualize class assignment to appropriate classifier. The pre- diction either indicates CNN superiority and appropriateness for usage with the input test sample. Otherwise, SVM shall be used. Thus, CNN is used when SVM performance is lower than CNN and vice versa. At the last stage, the untouched testing set is used for final assessment of the performance of the fused classifiers. Occasionally for some testing samples, the results of both classifiers votes are within the area of specialization of each of the classifiers, so confusion matrix statistical informa- tion is used to determine the accuracy of each vote. An equa- tion is used to measure confidence depending on the history of misclassifications for the classes chosen by each classifier. For each test sample I.

$$\int(i) = MisclassificationRateCNN - AccuracyRateCNN - MisclassificationRateSVM \succ$$
$$MisclassificationRateSVM - AccuracyRateSVM - MisclassificationRateCNN$$

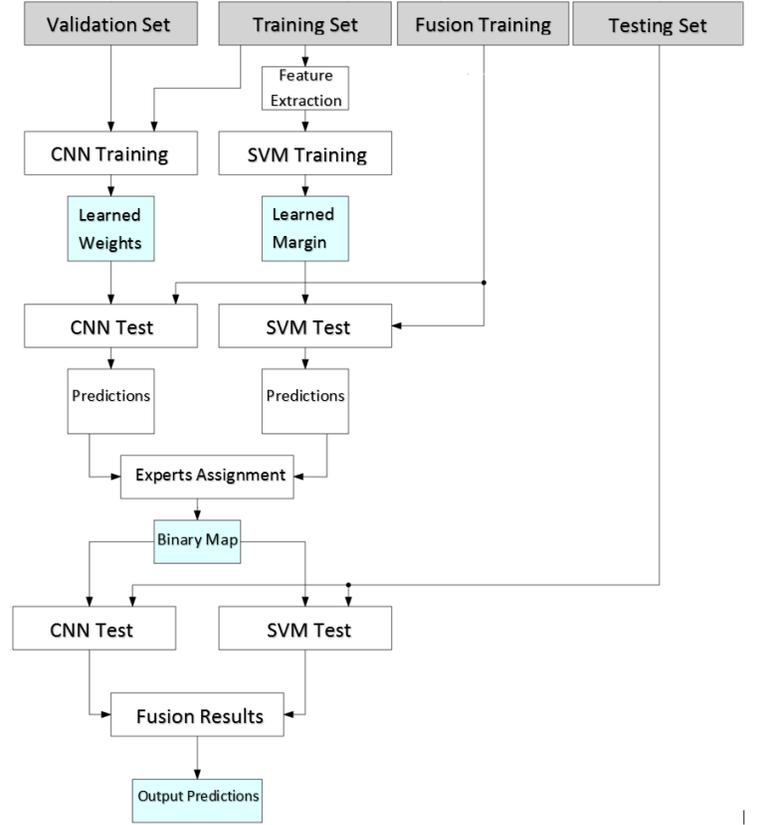

Fig. 2. Overall classification System Architecture

of particular class based on the error rate. The Misclassifica- tion Rate(A) is the percentage of invalid predictions of samples predicted to be a class A while they belong to any other class. The Accuracy Rate(A) is the percentage of valid predictions of samples predicted to be class A. The first segment of the left- hand side is the confidence in which any samples predicted to be a class A would turn out to be valid. The percentage of misclassifications predicted to be a class A committed by the other classifier is added a positive factor.

## V. EXPERIMENTS

The fusion binary mapping layer has one primary parameter that requires training. Thus, the fusion is affected by data distribution among training, and testing. Several experiments had been conducted to estimate the best allocation of data for each dataset. For both datasets, experimentation for the singled out parameter was done with a constant ratio of distribution to rest of the data among other parameters. The training data had been iteratively increased for both datasets to test the effect of training increase on the fused classifier performance. The rest of the parameters (Validation, Mapping, and test) got data in the ratio of 1-1-2 that is to give testing the highest priority. The constant rate of distribution attempt to freeze other parameters and focus on the training data percentage increase. For Brodatz dataset, the training parameter was tested and returned results shown in Table I.

| Training Percentage | Classification Accuracy |
|---|---|
| 10% | 74.66% |
| 20% | 79.79% |
| 30% | 90.62% |
| 40% | 92.70% |
| 50% | 85.74% |
| 60% | 93.35% |
| 70% | 93.75% |

TABLE I
TRAINING DATA PERCENTAGE ACCURACY FOR BRODATZ DATASET

| Training Percentage | Classification Accuracy |
|---|---|
| 10% | 95.98.% |
| 20% | 98.71%% |
| 30% | 98.80% |
| 40% | 98.54% |
| 50% | 99.25% |
| 60% | 99.55% |
| 70% | 98.88% |

TABLE II
TRAINING DATA PERCENTAGE ACCURACY FOR KYLBERG DATASET

It was concluded from the first experiment that around 60% of the data for training are sufficient for reasonable performance. Similar testing was conducted for Kylberg dataset and returned results shown in Table II. The marginally higher results for Kylberg dataset than for Brodatz at low training data is attributed to low similarity between different classes. Thus, the easier separation between its texture, unlike Brodatz that has texture types that look very similar. The second parameter tested is mapping or fusion training parameter. The testing involved the same methodology used for classifiers training test. The training data increased iteratively by 10 percent per iteration. The remaining data were distributed in the ratio of 2-1-1 on training, validation, testing and giving emphasis on classifier training to fusion training relation. The distribution was done under the assumption that poor classifier training will result in a bad fusion regardless of how proficient the fusion algorithm might appear to be.

The mapping parameter was tested for both classifiers as well. The results for Brodatz are shown in Table III.

The results show that the increase of training data would have a positive effect on fusion but taking into consideration that dedicating much of the data to fusion layer training would prohibit optimal training of CNN and SVM. Thus, an inversely proportional relation between the classifier training to fusion training. It is noticed that abrupt increase in fusion training data portion, leads to a big decrease in classifier training performance, causing a high instability in training. Moreover, the retraining of classifiers and number of iterations given the amount of data, which was constant in this case cause random increases, or decreases of classification results. The lower the data and a constant high number of epochs may lead to overfitting. The results of mapping parameter for Kylberg dataset are shown in Table IV.

| Training Percentage | Classification Accuracy |
|---|---|
| 10% | 90.62.% |
| 20% | 90.36%% |
| 30% | 87.23% |
| 40% | 89.45% |
| 50% | 86.71% |
| 60% | 90.62% |
| 70% | 89.84% |

TABLE III
BINARYMAP ACCURACY PER PERCENTAGE OF TRAINING DATA FOR BRODATZ DATASET

| Training Percentage | Classification Accuracy |
|---|---|
| 10% | 99.44.% |
| 20% | 99.44%% |
| 30% | 98.77% |
| 40% | 98.88% |
| 50% | 98.88% |
| 60% | 98.88% |
| 70% | 98.87% |

TABLE IV
BINARYMAP ACCURACY PER PERCENTAGE OF TRAINING DATA FOR KYLBERG DATASET

The mapping parameter wasn't profoundly affected by small training data due to the SVM efficiency of separation under low training data. In most of Kylberg dataset results, the SVM results were more stable and accurate than CNN. Thus, for the Kylberg dataset in specific, the fusion parameter amount of training data wasn't of a concern compared with training data. The adopted percentage of data distribution for Kylberg given the results obtained above was 60% for training data, 10% for validation, 20% for fusion, and 10% for testing. The distribution for Brodatz was 60% for training, 10% for validation, 20% for fusion, and 10% for testing which suggests that a generalized amount of training to testing data percentages had been reached.

The datasets have been fully shuffled for each test and averaged to obtain an average of performance. For Kylberg dataset, the SVM performed superiorly to CNN, and that is due to lack of any rotated or scaled objects in the dataset thus, the separation margin was more accurate with SVM.

|  | Exp. 1 | Exp. 2 | Exp. 3 | Average |
|---|---|---|---|---|
| CNN | 97.32% | 95.53% | 95.31% | 96.05% |
| SVM | 100% | 99.33% | 99.55% | 99.62% |
| Fusion | 100% | 99.33%% | 99.55% | 99.62% |

TABLE V
FUSION RESULTS FOR KYLBERG DATASET

The Brodatz dataset is perfect for testing both classifiers strength points where rotation and scaling are present in the dataset. For experiment 1 the accuracy rate was 98.43% for both classifiers, yet the misclassified samples were not the same samples for both classifiers which show the potential improvement accuracy rates based on the fusion method.

|  | Exp. 1 | Exp. 2 | Exp. 3 | Average |
|---|---|---|---|---|
| CNN | 98.43% | 97.65% | 91.40% | 95.82% |
| SVM | 98.43% | 92.96% | 94.53% | 95.30% |
| Fusion | 98.43% | 99.21% | 96.87% | 98.17% |

TABLE VI
FUSION RESULTS FOR BRODATZ DATASET

## VI. RESULTS

Throughout the experiments performed, the percentage of training set was the primary factor of classifier performance. At a particular threshold, additional training data wouldn't show significant major improvement in testing results. It was found that the fusion algorithm has reached 99.21%. On the other hand, the random initial weights play an important factor in achieving optimal results. Even with same training set, the results of CNN may be entirely different from the previous training attempts. On the other hand, SVM is relatively stable due to different mechanism employed for training and obtaining margin. It's worth noting that choosing such setup allowed for fault tolerance to take place when CNN training failed to recognize the considerable amount of testing set samples, SVM was able to take over and obtain acceptable fusion results. The CNN achieved a testing accuracy of 91.40% on the third experiment, which shows the effect of proper training on different samples of data and initial weights.

## VII. CONCLUSIONS

Both CNN and SVM have different learning approaches. Their fusion would usually lead to improvement in classification rates due to uncorrelated errors. In some cases, one classifier may be superior to the other, and in that case, fusion would still ensure a reasonable final classification result regardless of one of the classifiers lacking in accuracy due to nature of the problem. Thus, the proposed fused classifier it able to recover from mistakes of any it's classifiers and obtains high prediction under the condition that both classifiers are well trained.